\documentclass{article}

\usepackage[preprint]{manustyle}

\usepackage[utf8]{inputenc} 
\usepackage[T1]{fontenc}    
\usepackage{hyperref}       
\usepackage{url}            
\usepackage{booktabs}       
\usepackage{amsfonts}       
\usepackage{amsmath}  
\usepackage{nicefrac}       
\usepackage{microtype}      
\usepackage{xcolor}         
\usepackage{graphicx}
\usepackage{multirow}

\title{Hyperbolic Coarse-to-Fine Few-Shot Class-Incremental Learning}

\author{%
  Jiaxin Dai and Xiang Xiang\\ 
  HUST AI and Visual Learning Lab (HAIV Lab)\\
  Huazhong University of Science and Technology (HUST), China\\
  \texttt{xex@hust.edu.cn} \\
}

\begin{document}

\maketitle

\begin{abstract}
In the field of machine learning, hyperbolic space demonstrates superior representation capabilities for hierarchical data compared to conventional Euclidean space. This work focuses on the Coarse-To-Fine Few-Shot Class-Incremental Learning (C2FSCIL) task. Our study follows the Knowe approach, which contrastively learns coarse class labels and subsequently normalizes and freezes the classifier weights of learned fine classes in the embedding space. To better interpret the "coarse-to-fine" paradigm, we propose embedding the feature extractor into hyperbolic space. Specifically, we employ the Poincaré ball model of hyperbolic space, enabling the feature extractor to transform input images into feature vectors within the Poincaré ball instead of Euclidean space. We further introduce hyperbolic contrastive loss and hyperbolic fully-connected layers to facilitate model optimization and classification in hyperbolic space. Additionally, to enhance performance under few-shot conditions, we implement maximum entropy distribution in hyperbolic space to estimate the probability distribution of fine-class feature vectors. This allows generation of augmented features from the distribution to mitigate overfitting during training with limited samples. Experiments on C2FSCIL benchmarks show that our method effectively improves both coarse and fine class accuracies.
\end{abstract}


\section{Introduction}

In real-world computer vision applications, models must address three key challenges: dynamic environments, limited training data, and evolving knowledge. First, practical scenarios often involve open and changing conditions (\cite{FewShot01}). For example, autonomous driving systems must adapt to varying lighting conditions and seasonal changes, requiring models to update incrementally without forgetting previous knowledge. Second, many tasks face the problem of limited labeled data (\cite{FewShot02}). For example, wildlife monitoring may have only a few images of rare species, or industrial systems may lack sufficient training samples for new equipment models, making traditional data-heavy methods impractical. Most importantly, humans use hierarchical understanding when learning new concepts, first recognizing broad categories (e.g., "birds") before learning detailed features (e.g., feather patterns of specific rare species) with minimal examples. This coarse-to-fine approach efficiently combines existing knowledge with new information. However, current methods often treat these challenges separately: fixed category settings fail to handle new data, single models struggle to use both general and detailed features, and frequent full retraining is unrealistic for real-world deployment. Therefore, combining coarse-to-fine hierarchical learning, few-shot learning, and incremental learning is critical to building practical vision systems. In this paper,we are interested in such Coarse-to-Fine Few-Shot Class-Incremental Learning (C2FSCIL) problem, aiming to build sustainable learning systems that better align with real-world requirements.

We follow the classical Knowe algorithm (\cite{Knowe}), which is designed for C2FSCIL tasks by contrastively learning coarse class labels and subsequently normalizing and freezing the classifier weights of learned fine classes in the embedding space. Given that the learning task from coarse to fine classes can be viewed as a hierarchical transformation of data, and hierarchical data structures are typically more suitable for representation in hyperbolic space rather than Euclidean space, this study improves the Knowe algorithm as follows.

For the feature extractor, we train it in hyperbolic space to ensure the output feature vectors reside in hyperbolic space instead of Euclidean space. This is achieved by introducing a hyperbolic mapping layer after the feature extractor and conducting corresponding model training operations.

For the loss function, we adapt the original contrastive loss in the Knowe algorithm to hyperbolic space, reformulating it as a hyperbolic contrastive loss.

For the classifier, we replace the standard fully connected layer in the classifier with a hyperbolic version, enabling classification of hyperbolic feature vectors and leveraging hyperbolic space’s inherent strength in modeling hierarchical data.

Additionally, since the task involves few-shot learning, the study further refines the Knowe algorithm from this perspective. In experiments, we mitigate overfitting in few-shot learning by estimating the data distribution. Specifically, we estimate the mean and variance of the hyperbolic space distribution based on training sample feature vectors, then sample additional training examples from this estimated distribution to augment the training set for classifier optimization.

\section{Related Work}

\paragraph{Coarse-to-fine learning}
Hierarchical learning models the dependencies between categories through hierarchical label structures or breaks down classification tasks into a step-by-step process from coarse to fine levels. Existing methods mainly fall into three categories: network architecture-based methods (\cite{Capsule}), loss function-base methods (\cite{Discriminative}) and feature space-based methods (\cite{HypHierarchical}).

\paragraph{Class-incremental learning}
Incremental learning allows models to continuously update with new data without forgetting previous knowledge, instead of training on all data at once. Based on task settings, it can be divided into class-incremental learning and task-incremental learning. This paper focuses on class-incremental learning. Common methods for incremental learning mainly fall into four categories: regularization-based methods (\cite{LwF}), generative replay-based methods (\cite{iCarL}), parameter isolation-based methods (\cite{PackNet}), and knowledge distillation-based methods (\cite{LwM}).

\paragraph{Few-shot learning}
Few-shot learning enables models to acquire useful knowledge with very limited samples, helping address the problem of imbalanced sample sizes between different classes. Common few-shot learning approaches can be categorized into three main types: fine-tuning-based methods (\cite{Finetuning}), data augmentation-based methods (\cite{Aug01,Aug02,Aug03}), and transfer learning-based methods (\cite{Transfer01,Transfer02,Transfer03}).

\section{Preliminaries}
\label{hyperbolic}

{\bf Normal Space:}
By the central limit theorem and maximum entropy principle, machine learning problems in Euclidean space often assume samples follow a normal distribution. However, in general Riemannian spaces, the limiting distribution from the central limit theorem and the maximum entropy distribution are typically distinct. Due to the complexity of the central limit theorem in hyperbolic space, this paper assumes that data distributions satisfy the maximum entropy distribution—i.e., the distribution with maximum entropy given the mean and variance, which also implies the most uniform encoding of unknown information and minimal assumptions about data morphology.
According to Riemannian geometry research (\cite{Intrinsic}), the probability density function of the maximum entropy distribution in Riemannian space is given by:
\[{\cal N}({\bf{x}};\mu ,\Gamma ) = k\exp \left( { - \frac{1}{2}{{({{\log }_\mu }{\bf{x}})}^T}\Gamma ({{\log }_\mu }{\bf{x}})} \right)\]
where $\mu$ is the mean, $\Gamma$ is the precision matrix, ${\log _\mu }\left(  \cdot  \right)$ is the logarithmic map, and $k$ is the normalization constant. In Euclidean space, $\Gamma$ is the inverse of the covariance matrix, but this relationship does not hold in general Riemannian spaces. For example, when $\Gamma$ is the Riemannian metric matrix $G\left( \mu  \right)$, the covariance matrix becomes the identity matrix, leading to
\begin{align*}
{\cal N}({\bf{x}};\mu ,\Gamma ) & = k\exp \left( { - \frac{1}{2}{{({{\log }_\mu }{\bf{x}})}^T}G\left( \mu  \right)({{\log }_\mu }{\bf{x}})} \right)\\
& = k\exp \left( { - \frac{1}{2}{{\left\| {{{\log }_\mu }{\bf{x}}} \right\|}_\mu }} \right)\\
& = k\exp \left( { - \frac{1}{2}{\rm{dist}}{{\left( {\mu ,{\bf{x}}} \right)}^2}} \right)
\end{align*}
For the Poincaré ball model, the Riemannian metric $G\left( \mu  \right) = {\left( {\lambda _\mu ^c} \right)^2}{I_n}$ is a constant multiple of the identity matrix. Thus, the probability density function of the maximum entropy distribution in the Poincaré ball model simplifies to
\[\mathcal{N}^c_{\mathbb{H}} (\mathbf{x} ; \mu, \Sigma) = k \exp \left( -
\frac{\mathbf{z}^T \Sigma^{- 1}  \mathbf{z}}{2} \right), \mathbf{z} =
\lambda_{\mu}^c \log_{\mu}^c  \mathbf{x}\]
where $\Sigma$ is the covariance matrix.

\begin{figure}[htbp]
  \centering
  \label{distri}
  \includegraphics[width=1\textwidth]{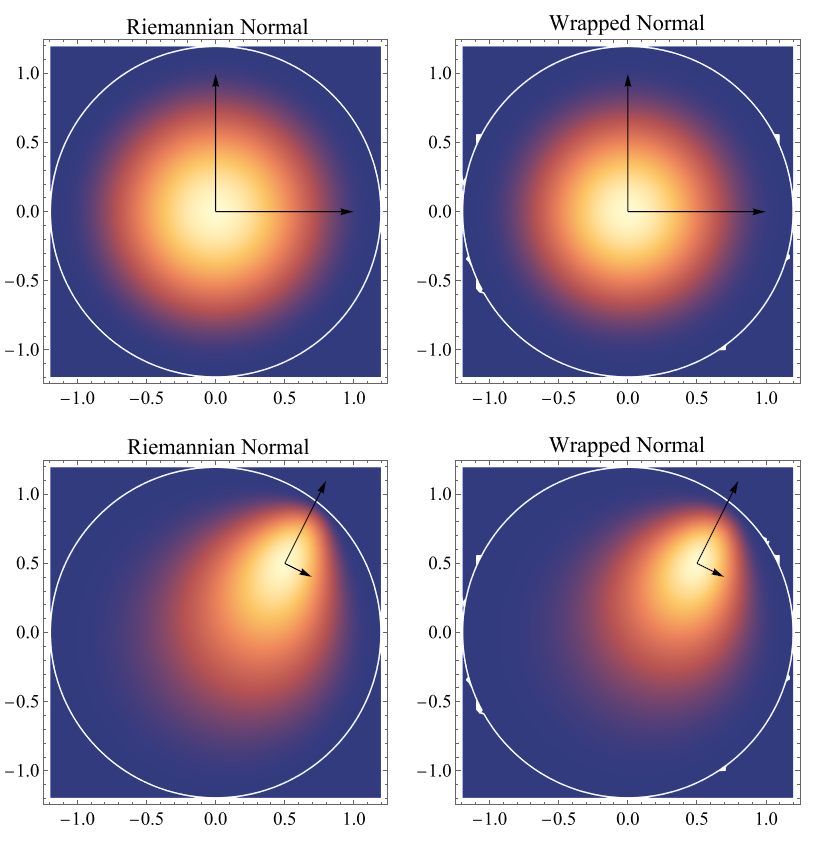}
  \caption{Riemannian normal distribution (maximum entropy distribution, left column) and wrapped normal distribution (right column) in hyperbolic space. Arrows represent the eigenvectors of the covariance matrix, with their origins indicating the mean.}
\end{figure}

However, the normalization constant of this distribution is computationally intractable, and uniform sampling is challenging. Therefore, in practical modeling, the wrapped normal distribution ${\cal N}_W^c({\bf{x}};\mu ,\Sigma )$ is often adopted (\cite{Wrapped}).Namely, Samples are first drawn from a standard normal distribution ${\cal N}(0,\Sigma )$ to obtain $v$, and then ${\bf{x}} = \frac{{\exp _\mu ^c{\bf{v}}}}{{\lambda _\mu ^c}}$ is computed as the sampling result of ${\cal N}_W^c({\bf{x}};\mu ,\Sigma )$. The actual maximum entropy distribution mentioned above is termed the Riemannian normal distribution. In practice, the probability density functions of the two distributions satisfy the following relationship (\cite{Variational}):
\[\mathcal{N}_W^c  (\mathbf{x} ; \mu, \Sigma) =\mathcal{N}_{\mathbb{H}}^c
(\mathbf{x} ; \mu, \Sigma) \left( \frac{\sqrt{c} \cdot d_p^c (\mu,
\mathbf{x})}{\sinh \left( \sqrt{c} \cdot d_p^c (\mu, \mathbf{x}) \right)}
\right)^{d - 1}\]
Since $\frac{{\sinh x}}{x} > 1$ and increases exponentially with $x$, the wrapped normal distribution includes an additional exponentially decaying term compared to the Riemannian normal distribution at larger distances from the mean. This indicates a lower probability of samples appearing far from the mean. Overall, for the same covariance, the wrapped normal distribution is more concentrated near the mean (see \ref{distri}) .

{\bf Hyperbolic Space:}
Machine learning tasks often involve mean computation, such as prototype extraction in prototype networks, feature standardization in batch normalization, and parameter estimation for distribution means based on samples.

In Euclidean space, computing the mean for a sample set is straightforward. For a sample set $X = \left( {{x_1},{x_2}, \ldots ,{x_n}} \right)$, where ${x_i} \in {\mathbb{R}^d}$, the mean $\overline X$ is the arithmetic average: $\overline X  = \frac{1}{n}\sum\limits_{i = 1}^n {{x_i}}.$

However, in hyperbolic space, due to the curvature-induced distortion of geodesics, the mean cannot be directly computed. For a sample set $X = \left( {{x_1},{x_2}, \ldots ,{x_n}} \right)$ in hyperbolic space, where ${x_i} \in \mathcal{B}_c^d$, two distinct definitions of the mean exist: the geometric mean and the statistical mean.

\begin{figure}[htbp]
  \centering
  \label{gmean}
  \includegraphics[width=1\textwidth]{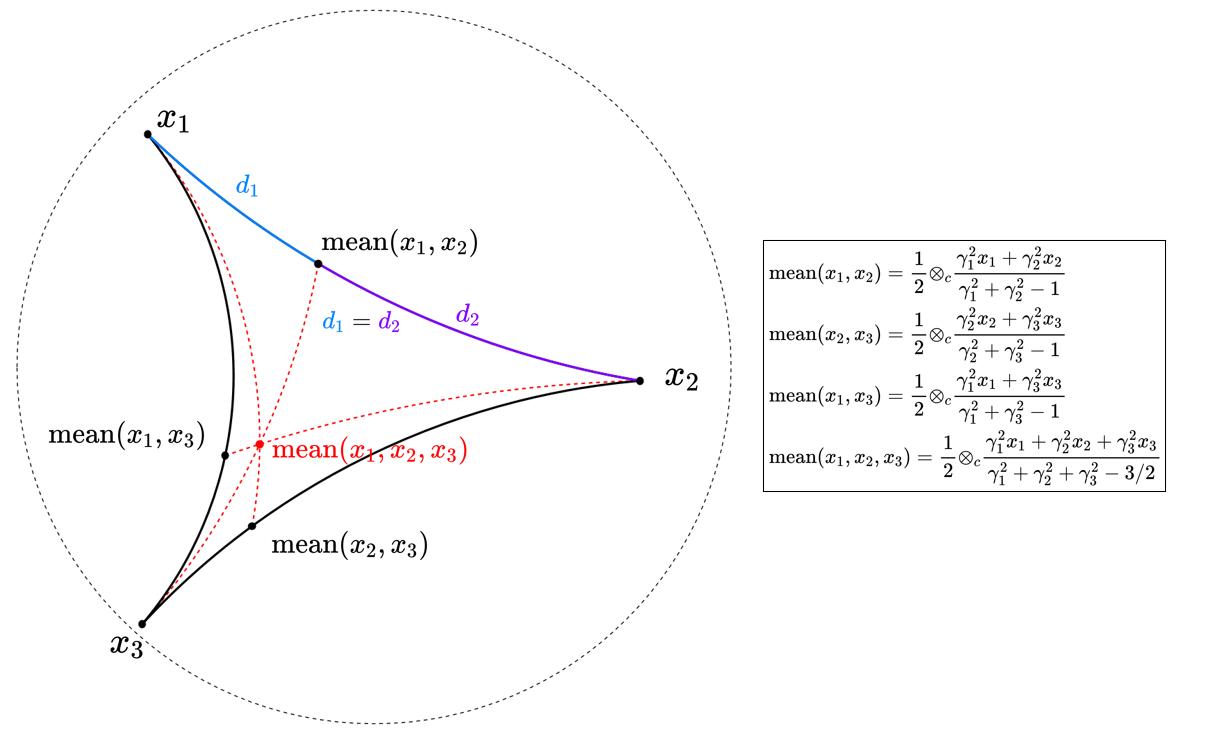}
  \caption{Geometric mean in hyperbolic space, where the outermost circle represents the boundary of the Poincaré ball. All sample points remain confined within this circle.}
\end{figure}

\emph{Estimating geometric mean}.
Also known as the Einstein midpoint or gyrocentroid, the geometric mean can be viewed as an analog of vector addition in Euclidean space within hyperbolic space (\cite{Analytic}). It assigns a weight to each vector based on its norm in hyperbolic space and computes the weighted average as the mean. The formula is:
\[\overline X  = \frac{1}{2}{ \otimes _c}\frac{{\sum\limits_{i = 1}^n {2\gamma _i^2{x_i}} }}{{\sum\limits_{i = 1}^n {\left( {2\gamma _i^2 - 1} \right)} }},\]
where ${\gamma _i} = \frac{1}{{\sqrt {1 - c{{\left\| {{x_i}} \right\|}^2}} }}$ is the Lorentz factor, and ${ \otimes _c}$ represents a degenerated form of matrix multiplication to scalar multiplication, defined as:
\[r{ \otimes _c}{\mathbf{x}} = \frac{1}{{\sqrt c }}\tanh \left( {r{{\tanh }^{ - 1}}\left( {\sqrt c \left\| {\mathbf{x}} \right\|} \right)} \right)\frac{{\mathbf{x}}}{{\left\| {\mathbf{x}} \right\|}}\]
Notably, for two points, the geometric mean is equidistant from both in hyperbolic space. For three points, it is the intersection of the three geodesics connecting each point to the geometric mean of the other two, analogous to the centroid of a triangle in Euclidean space (see \ref{gmean}).

\emph{Estimating statistical mean}. Also known as the conformal barycenter or Fréchet mean, the statistical mean is defined as the point that minimizes the sum of squared distances to all sample points:
\[\overline X  = \mathop {\arg \min }\limits_{x \in \mathcal{B}_c^d} \sum\limits_{i = 1}^n {{\text{dist}}{{\left( {x,{x_i}} \right)}^2}} \]
In Euclidean space,${\text{dist}}\left( {{\mathbf{x}},{\mathbf{y}}} \right) = \left\| {{\mathbf{x}} - {\mathbf{y}}} \right\|$, and the statistical mean reduces to the arithmetic mean of all samples. However, in hyperbolic space, ${\text{dist}}\left( {{\mathbf{x}},{\mathbf{y}}} \right) = d_p^c({\mathbf{x}},{\mathbf{y}})$, and for general curvature $c$, sample size $n$, and dimensionality $d$, the statistical mean lacks a closed-form solution but can be solved via gradient descent. Specifically, let the initial estimate ${\mu _0}$ of the statistical mean $\mu$ be the geometric mean of the sample set. Then iteratively update using:
\[{\mu _{t + 1}} = \exp _{{\mu _t}}^c\left( {\frac{1}{n}\sum\limits_{i = 1}^n {\log _{{\mu _t}}^c\left( {{x_i}} \right)} } \right),\]
until ${\mu _t}$ converges to the statistical mean. Notably, when $n = 2$, the statistical mean coincides with the geometric mean, both being the point equidistant from the two samples.

For a probability distribution in hyperbolic space with mean $\mu$, drawing sufficiently many samples will yield a statistical mean tending towards $\mu$.Similarly, for a sample set, if we wish to estimate the probability distribution that the samples follow (e.g., during standardization), the statistical mean should also be used as the estimator for the distribution mean. However, for a large sample set or when the sample dimensionality is high, the computational cost of calculating the statistical mean is substantial. Therefore, this paper employs the Aaron algorithm  (\cite{Differentiating}) to compute the statistical mean. This method is specifically designed for solving statistical means in the Poincaré ball model and offers advantages of faster convergence and stability compared to the gradient descent method mentioned above.

\emph{Estimating variance}.
Under the few-shot condition, where the number of training samples is far smaller than the dimensionality of the feature vectors, the covariance matrix $ \Sigma $ estimated from the training samples will contain a large number of zero eigenvalues. This confines the probability distribution to a low-dimensional hyperplane determined by the training samples, which is meaningless. Therefore, in practical models, it is assumed that $\Sigma  = {\sigma ^2}{I_n}$, where \[ {\sigma ^2} = \frac{1}{n}\sum\limits_{k = 1}^n {d_p^c{{\left( {\mu ,{x_k}} \right)}^2}} \] is the variance.

\section{Proposed Methodology}

\subsection{Learning Hyperbolic Network}

{\bf Hyperbolic mapping layer:}
The purpose of the hyperbolic mapping layer ${\text{TP}}\left(  \cdot  \right)$ is to map feature vectors from Euclidean space to hyperbolic space. To achieve this, the original Euclidean feature space can be regarded as the tangent space at point ${\mathbf{w}}$ within the Poincaré ball, and the mapping from Euclidean space to hyperbolic space is realized through the exponential map $\exp _{\mathbf{w}}^c$. Specifically, for an input feature vector ${\mathbf{x}} \in {\mathbb{R}^n}$, the hyperbolic mapping layer is defined as ${\text{TP}}\left( {\mathbf{x}} \right){\text{ = }}\exp _{\mathbf{w}}^c({\mathbf{x}})$, where 
${\mathbf{w}} \in \mathcal{B}_c^n$ can be considered as the weight of this layer. Both $w$ and the curvature $c$ are learnable parameters that can be updated via gradient descent. However, for convergence considerations, $c$ is generally not directly learned here but is predefined and frozen.

\begin{figure}[t!]
  \centering
  \includegraphics[width=0.9 \textwidth]{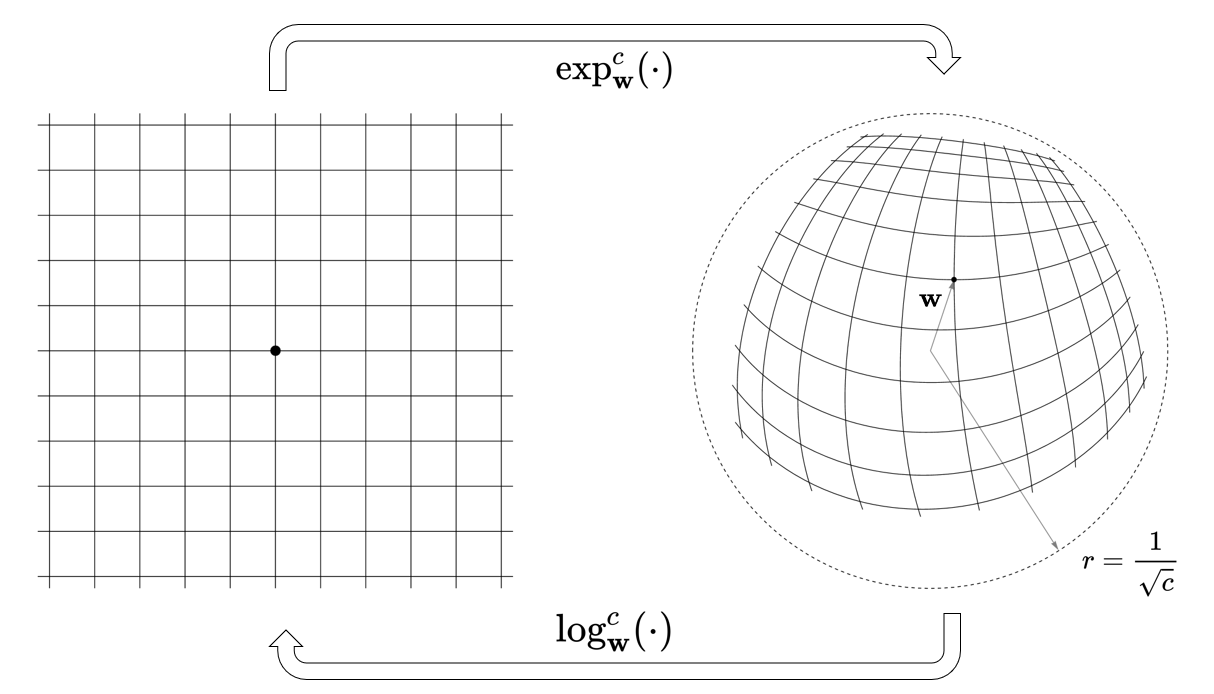}
  \caption{A schematic diagram of hyperbolic space mapping. The left diagram represents Euclidean space, the right diagram represents hyperbolic space, and the two are interconverted through the exponential map and logarithmic map.}
\end{figure}

{\bf Hyperbolic fully-connected layer:}
For an input vector ${\mathbf{x}} \in {\mathbb{R}^n}$ , the fully connected (FC) layer in Euclidean space performs a linear transformation, i.e., ${\mathbf{y}} = {\text{FC}}\left( {\mathbf{x}} \right) = W{\mathbf{x}} + b$, where $ W \in {\mathbb{R}^{n \times m}}$ is the weight matrix, $b \in {\mathbb{R}^m}$ is the bias term, and ${\mathbf{y}} \in {\mathbb{R}^m}$ is the output vector. Matrix multiplication can essentially be viewed as a linear mapping $f:{\mathbb{R}^n} \to {\mathbb{R}^m}$ . For the Euclidean mapping $f$ , its counterpart in hyperbolic space is $ {f_\mathbb{H}}\left( {\mathbf{x}} \right) = \exp _0^c\left( {f\left( {\log _0^c\left( {\mathbf{x}} \right)} \right)} \right)$, which involves first transforming the input vector into Euclidean space via the logarithmic map, applying the mapping $f$, and then mapping it back to hyperbolic space. Since matrix multiplication corresponds to the mapping 
${\mathbf{x}} \to M{\mathbf{x}}$, its hyperbolic counterpart is defined as:\[M{ \otimes _c}{\mathbf{x}} = \frac{1}{{\sqrt c }}\tanh \left( {\frac{{\left\| {Mx} \right\|}}{{\left\| x \right\|}}{{\tanh }^{ - 1}}\left( {\sqrt c \left\| x \right\|} \right)} \right)\frac{{Mx}}{{\left\| {Mx} \right\|}},\] thus defining the hyperbolic fully connected layer ${\text{HypFC}}\left(  \cdot  \right)$: \[{\mathbf{y}} = {\text{HypFC}}\left( {\mathbf{x}} \right) = W{ \otimes _c}{\mathbf{x}}{ \oplus _c}b,\]where the weight matrix $ W \in {\mathbb{R}^{n \times m}}$, the bias term $b \in \mathcal{B}_c^n$, and the curvature $c$ are all network parameters of this layer (\cite{HNN}) .

{\bf Hyperbolic contrastive loss:}
Contrastive loss can be used to capture intra-class fine-grained differences and plays a crucial role in this study. The contrastive loss in Euclidean space is formulated as (\cite{MoCo}):\[{\mathcal{L}_{Con}} =  - \mathop {\mathop \sum \limits^N }\limits_{n = 1} \log \frac{{\exp ({\mathbf{q}}_n^T{\mathbf{k}}_n^ + /\tau )}}{{\exp ({\mathbf{q}}_n^T{\mathbf{k}}_n^ + /\tau ) + \mathop \sum \limits_{m \ne n} \exp ({\mathbf{q}}_n^T{\mathbf{k}}_m^ - /\tau )}},\]However, in hyperbolic space, due to the curvature-induced distortion of geodesics, the inner product between vectors differs from that in Euclidean space. Its analogy in hyperbolic space should achieve an effect similar to the inner product, i.e., the value increases when two vectors are closer and decreases when they are farther apart. This paper replaces the inner product term in the original loss function with the negative hyperbolic distance (\cite{Contrastive}), i.e.,\[\mathcal{L}_{Con}^{hyp} =  - \mathop {\mathop \sum \limits^N }\limits_{n = 1} \log \frac{{\exp \left[ { - d_p^c\left( {{{\mathbf{q}}_n},{\mathbf{k}}_n^ + } \right)/\tau } \right]}}{{\exp \left[ { - d_p^c\left( {{{\mathbf{q}}_n},{\mathbf{k}}_n^ + } \right)/\tau } \right] + \mathop \sum \limits_{m \ne n} \exp \left[ { - d_p^c\left( {{{\mathbf{q}}_n},{\mathbf{k}}_m^ - } \right)/\tau } \right]}}.\]

\subsection{Hyperbolic Few-Shot Class-Incremental Learning for Coarse-To-Fine Recognition}

Our method HypKnowe closely follows Knowe: it first learns coarse-class features via contrastive learning, then learns, normalizes, and freezes fine-class features. During coarse-class learning, the algorithm acquires feature embedding weights using a contrastive learning framework similar to ANCOR, with MoCo as the backbone network. The final fully connected layers of the two data streams are replaced with MLPs. Additionally, HypKnowe introduces a hyperbolic mapping layer ${\rm{TP}}\left(  \cdot  \right)$ before the MLPs. The MLP hidden layers of the two streams output intermediate features ${\bf{q}}$ and ${\bf{k}}$. For given coarse labels, the total loss is defined as ${{\cal L}^C} = {\cal L}_{{\rm{Con}}}^{{\rm{Hyp}}} + {\cal L}_{CE}^C,$ where 
\[{\cal L}_{\rm{Con}}^{\rm{Hyp}} =  - \mathop {\mathop \sum \limits^N }\limits_{n = 1} \log \frac{{\exp \left[ { - d_p^c\left( {{{\bf{q}}_n},{\bf{k}}_n^ + } \right)/\tau } \right]}}{{\exp \left[ { - d_p^c\left( {{{\bf{q}}_n},{\bf{k}}_n^ + } \right)/\tau } \right] + \mathop \sum \limits_{m \ne n} \exp \left[ { - d_p^c\left( {{{\bf{q}}_n},{\bf{k}}_m^ - } \right)/\tau } \right]}}\]
is the hyperbolic contrastive loss, and ${\cal L}_{CE}^C$ is the standard cross-entropy loss to capture inter-class differences. Here, $N$ is the total number of samples, $\tau$ is the temperature parameter, ${\bf{k}}_m^ - $ denotes the intermediate layer output of the $m$-th sample (a negative sample),which belongs to the same coarse class as the $n$-th sample (a positive sample). This setup captures intra-class fine-grained differences while reducing interference for subsequent fine-grained classification, thereby improving accuracy. The hyperbolic contrastive loss ${\cal L}_{{\rm{Con}}}^{{\rm{Hyp}}}$ decreases when ${{\bf{q}}_n}$ is close to ${\bf{k}}_n^ + $ but distant from ${\bf{k}}_m^ - $.

\begin{figure}[htbp]
  \centering
  \includegraphics[width= \textwidth]{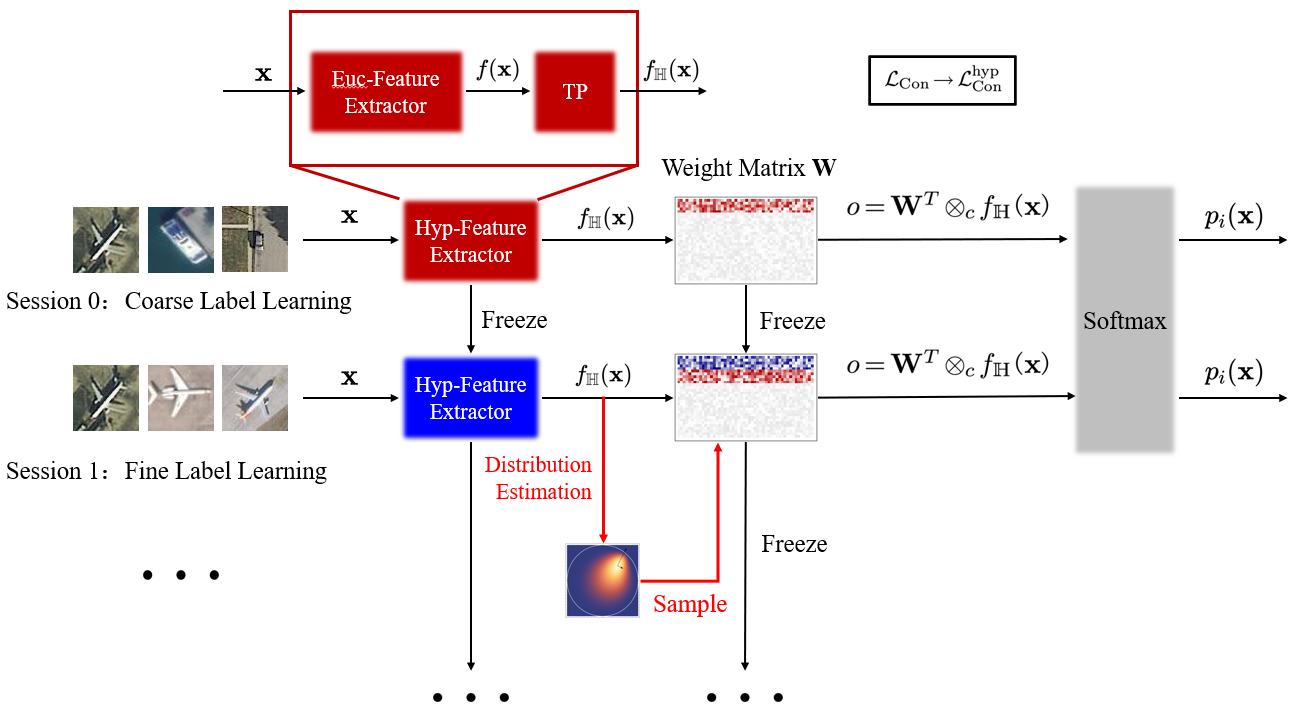}
  \caption{Schematic diagram of the HypKnowe algorithm, where $\mathcal{L}^{\text{Hyp}}_{\text{Con}}$ denotes the hyperbolic contrastive loss, $\otimes_c$ represents the operation of the hyperbolic fully connected layer, and $f_{\mathbb{H}} (\mathbf{x})$ denotes the feature vector in hyperbolic space.}
\end{figure}

In the incremental sessions, the objective of HypKnowe is to learn a weight matrix ${\bf{W}}$ with dimensions $D \times C$, where $D$ is the dimensionality of the feature vector output by the feature extractor, and $C$ is the total number of classes, expressed as $C = {C_{coarse}} + {C_{fine}}$ , where $C_{coarse}$ is the number of coarse classes and $C_{fine}$ is the number of fine classes. Assuming the feature extractor learned on coarse class labels is denoted as $f_{\mathbb{H}} (\cdot)$, for an input sample ${\bf{x}}$, it outputs a $D$-dimensional feature vector $f_{\mathbb{H}} (\mathbf{x})$. The network then computes the $C$-dimensional logits $o =\mathbf{W}^T \otimes_c f_{\mathbb{H}} (\mathbf{x})$ , where ${ \otimes _c}$ represents the matrix multiplication in the hyperbolic fully connected layer.

Subsequently, HypKnowe converts the logits into probability values via the Softmax activation function to classify ${\bf{x}}$. Specifically, the probability ${p_i}\left( {\bf{x}} \right)$ that the sample belongs to class $i$ is
\[p_i (\mathbf{x}) = \mathrm{Softmax} \left( \frac{o}{\lambda} \right) =
\frac{\exp (\mathbf{W}_i^T \otimes_c f_{\mathbb{H}} (\mathbf{x}) /
\lambda)}{\sum_j \exp (\mathbf{W}_j^T \otimes_c f_{\mathbb{H}} (\mathbf{x})
/ \lambda)},\]
where ${{\bf{W}}_j}$ denotes the $j$-th column of the weight matrix (i.e., the weight vector corresponding to class $j$), and $\lambda$ is the temperature parameter. During the incremental session, the loss function for optimization is the cross-entropy loss:
\[{\cal L}_{CE}^{\left( t \right)} =  - \frac{1}{{{C_t} \cdot K}}\sum\limits_{n = 1}^{{C_t} \cdot K} {\sum\limits_{i = 1}^{{N_t}} {\left[\kern-0.15em\left[ {y_n^{\left( t \right)} = i} 
 \right]\kern-0.15em\right]\log \left[ {{p_i}\left( {{\bf{x}}_n^{\left( t \right)}} \right)} \right]} }, \]
where ${C_t}$ is the number of classes learned at session $t$, $K$ is the number of samples per class in few-shot learning, $N_t$ is the total number of learned classes, and $\left[\kern-0.15em\left[  \cdot  
 \right]\kern-0.15em\right]$ is an indicator function that equals 1 if the expression is true and 0 otherwise.

Normalization refers to normalizing both the weight matrix and feature vectors, then computing normalized logits. Specifically, during logits computation, let
\[\widetilde{\mathbf{\mathbf{W}}_i} = \frac{\mathbf{\mathbf{W}}_i}{|
\mathbf{\mathbf{W}}_i |}, \widetilde{f_{\mathbb{H}}} (\mathbf{x}) =
\frac{f_{\mathbb{H}} (\mathbf{\mathbf{x}})}{d_p^c (0, f_{\mathbb{H}}
(\mathbf{\mathbf{x}}))},\]
resulting in normalized logits $\tilde{o} = \widetilde{\mathbf{\mathbf{W}}}^T \otimes_c
\widetilde{f_{\mathbb{H}}} (\mathbf{\mathbf{x}})$.Normalized weights and features are also used in probability computation to avoid classifier bias towards new classes.

Freezing implies that during incremental sessions, parameters of the feature extractor trained on coarse class labels are frozen, and the algorithm only processes its output feature vectors. Additionally, previously learned weight matrix parameters are frozen. For example, at session $t$, the model learns classes numbered from $C_t$ to $C_{t+1}$, where $t=0$ corresponds to the base session (learning coarse features, $C_0=0,C_1=C_{coarse}$). Columns $C_t$ to $C_{t+1}$ of $\mathbf{W}$ are updated, while columns prior to $C_t$ remain unchanged.

Additionally, HypKnowe introduces data augmentation via distribution estimation. In each incremental session, for every fine class learned in that session, the feature vectors ${v_i}$ of all training samples under the class are extracted using the feature extractor. Based on the aforementioned principles, the mean $\mu$ and variance ${\sigma ^2}$ of all feature vectors are computed. Then, augmented samples ${u_i}$ are drawn from the wrapped normal distribution 
${\cal N}_W^c({\bf{x}};\mu ,{\sigma ^2}I)$, labeled with the corresponding class, and fed into the classifier alongside the original feature vectors ${v_i}$ to assist in updating the weights.

\section{Experiments}

\subsection{Datasets}
\paragraph{CIFAR-100} The CIFAR-100 dataset (\cite{CIFAR}) contains 60,000 images from 100 fine classes, with each fine class comprising 500 training images and 100 test images. These fine classes are grouped into 20 coarse classes, each containing 5 fine classes. For example, the "trees" coarse class includes fine classes such as maple, oak, pine, palm, and willow.
\paragraph{tieredImageNet} The tieredImageNet dataset (\cite{tieredImageNet}) is a subset of ImageNet, containing 608 classes divided into 34 high-level super-classes to ensure significant semantic differences between training and test categories. The training/validation/test sets include 20, 6, and 8 coarse classes, 351, 97, and 160 fine classes, and 448K, 124K, and 206K images, respectively.
\paragraph{OpenEarthSensing} The OpenEarthSensing(OES) dataset (\cite{OES}) contains 157,674 remote sensing images from 189 fine classes, with each fine class containing hundreds to thousands of images. These fine classes are grouped into 10 coarse classes. For instance, the "aviation" coarse class includes fine classes such as A220, A330, Boeing 737, Boeing 747, and ARJ21.

\subsection{Performance Metrics}

In the experiments, model performance is evaluated based on four metrics: coarse-class accuracy, fine-class accuracy, average accuracy, and forgetting rate. At the end of each session, the algorithm separately computes the classification accuracy on coarse classes ${\cal A}_{{\rm{coarse}}}^n$ , fine classes ${\cal A}_{{\rm{fine}}}^n$ , and total accuracy ${\cal A}_{{\rm{total}}}^n$ , where $n$ denotes the index of the current session.

For $n = 0$ (the coarse-class learning session), ${\cal A}_{{\rm{coarse}}}^0$ represents the $\textbf{coarse-class accuracy}$. Since no fine-class labels are learned at this session, ${\cal A}_{{\rm{fine}}}^0 = 0$ and ${\cal A}_{{\rm{total}}}^0 = {\cal A}_{{\rm{coarse}}}^0$. For the last session $n = T$, ${\cal A}_{{\rm{fine}}}^T$ represents the $\textbf{fine-class accuracy}$.

In subsequent incremental sessions, the $\textbf{average accuracy}$ $\overline {\cal A} $ is calculated as the mean of total accuracies across all previous sessions:
\[\overline {\cal A}  = \frac{1}{{n + 1}}\sum\limits_{k = 0}^n {{\cal A}_{{\rm{total}}}^k} \]

The fine-class forgetting rate and coarse-class forgetting rate are defined as:
\[{\cal F}_{{\rm{fine}}}^n = \frac{{{\cal A}_{{\rm{fine}}}^{n - 1} - {\cal A}_{{\rm{fine}}}^n}}{{{\cal A}_{{\rm{fine}}}^{n - 1}}},{\cal F}_{{\rm{coarse}}}^n = \frac{{{\cal A}_{{\rm{coarse}}}^0 - {\cal A}_{{\rm{coarse}}}^n}}{{{\cal A}_{{\rm{coarse}}}^0}}\]
The overall $\textbf{forgetting rate}$ ${\cal F}$ is then computed as:
\[{\cal F} = \frac{1}{{T - 1}}\mathop {\mathop \sum \limits^T }\limits_{n = 2} {\cal F}_{{\rm{fine}}}^n \cdot \frac{{{C_n}}}{{{N_{{\rm{fine}}}}}} + \mathop {\mathop \sum \limits^{T - 1} }\limits_{n = 1} {\cal F}_{{\rm{coarse}}}^n \cdot \left( {1 - \frac{{{C_n}}}{{{N_{{\rm{fine}}}}}}} \right)\]
where $T$ is the total number of sessions, $C_n$ denotes the cumulative number of fine classes learned up to session $n$, and $N_{\rm{fine}}$ is the total number of fine classes.

\subsection{Implementation Details}
For the CIFAR-100 and OES datasets, the backbone network is ResNet12, while for the tieredImageNet dataset, the backbone network is ResNet50. During training, the optimization algorithm employs stochastic gradient descent (SGD) with a momentum parameter of 0.9, weight decay set to 5e-4, contrastive loss temperature parameter $\tau = 0.2$, and Softmax layer temperature parameter $\lambda = 0.5$. The batch size is 256 for tieredImageNet and CIFAR-100, and 64 for the OES dataset.

In the coarse-class learning session, the model is trained using the hyperbolic-enhanced ANCOR method (\cite{ANCOR}) with a learning rate of 0.12 for 200 epochs. The fine-class learning session involves few-shot tasks, with a learning rate of 0.1. The number of incremental sessions, new classes per session (Way), and training samples per new class (Shot) vary across datasets (see \ref{Para}). The number of query samples (Query) is fixed at 15 for all datasets.

\begin{table}[h]
\caption{\textbf{Parameter Settings for Coarse-To-Fine Few-Shot Incremental Learning Tasks}}
\label{Para}
\centering
\begin{tabular}{ccccccc}
\toprule
Dataset & Coarse Classes & Fine Classes & Sessions & Way & Shot & Query \\
\midrule
CIFAR-100 & 20 & 100 & 11 & 10 & 5 & 15 \\
tieredImageNet & 20 & 351 & 11 & 36 & 5 & 15 \\
OES & 10 & 189 & 13 & 15 & 5 & 15 \\
\bottomrule
\end{tabular}
\end{table}

For the hyperbolic space model, the curvature $c$ is set to the following empirical value:
\[c = {\left( {\frac{{\Gamma \left( {d/2 + 1} \right)}}{{{\pi ^{d/2 - 1}}}}} \right)^{ - 1/d}},\]
where $\Gamma \left(  \cdot  \right)$ denotes the gamma function, and $d$ is the dimensionality of the feature vectors. This curvature ensures that the volume (measure) of the Poincaré ball in $d$-dimensional space remains constant at $\pi$. For ResNet12 (output feature dimension $d=640$), the curvature is $c=0.162$, while for ResNet50 ($d=2048$), $c=0.091$. Throughout training, $c$ is frozen as a hyperparameter rather than a learnable parameter. The number of augmented samples generated via distribution estimation is set to 3.

\subsection{Ablation Study}

All experiments were conducted on a computational cluster equipped with four NVIDIA RTX 3090 GPUs. For the three datasets, the coarse-class pre-training phase (200 epochs) required approximately two days of total runtime. Subsequent fine-class few-shot learning sessions, based on the pre-trained model, were completed in approximately 30 seconds per session. Experiment results are shown in Table \ref{exr}.

\begin{table}[htbp]
\centering
\caption {Ablation Study Results for HypKnowe}
\label{exr}
\begin {tabular}{lcccccc}
\toprule
Dataset & Hyperbolic & Augmentation & ${\cal A}_{{\rm{coarse}}}^0$ & ${\cal A}_{{\rm{fine}}}^T$ & $\overline {\cal A} $ & ${\cal F}$ \\
\midrule
\multirow {3}{*}{CIFAR-100}
& $\times$ & $\times$ & 74.060 & 44.667 & 46.406 & 0.422 \\
& $\surd$ & $\times$ & \textbf{74.990} & 45.200 & 48.875 & \textbf{0.420} \\
& $\surd$ & $\surd$ & \textbf{74.990} & \textbf{46.400} & \textbf{50.528} & 0.421 \\
\midrule
\midrule
\multirow {3}{*}{tieredImageNet}
& $\times$ & $\times$ & 76.186 & 32.204 & 33.590 & \textbf{0.393} \\
& $\surd$ & $\times$ & \textbf{77.141} & 32.875 & 37.351 & 0.394 \\
& $\surd$ & $\surd$ & \textbf{77.141} & \textbf{33.107} & \textbf{37.822} & 0.394 \\
\midrule
\midrule
\multirow {3}{*}{OES}
& $\times$ & $\times$ & 80.734 & 49.043 & 51.066 & 0.250 \\
& $\surd$ & $\times$ & \textbf{81.011} & \textbf{51.879} & 58.848 & 0.202 \\
& $\surd$ & $\surd$ & \textbf{81.011} & 51.454 & \textbf{59.084} & \textbf{0.200} \\
\bottomrule
\end{tabular}
\end{table}

As shown in Table \ref{exr}, the introduction of hyperbolic space mechanisms simultaneously improves the model’s coarse-class accuracy, fine-class accuracy, and average accuracy. Further incorporating the data augmentation strategy based on probability distribution estimation enhances model performance to a greater extent, albeit with relatively limited improvement. This phenomenon can be attributed to the deviation between prototypes estimated via probability distribution under few-shot conditions and the actual prototypes, resulting in significant discrepancies between generated samples and real samples, thereby constraining the extent of improvement in the model’s generalization performance. Additionally, it should be noted that none of the aforementioned improvement strategies are optimized for incremental learning. Consequently, the model’s forgetting rate remains largely unchanged, which, on the other hand, validates that the proposed enhancements retain the original algorithm’s capability to mitigate catastrophic forgetting.

\section{Conclusion}

In this paper, we address the coarse-to-fine few-shot incremental learning task by proposing enhancements to the traditional Knowe method through the integration of hyperbolic space embedding and probability distribution estimation. Specifically, to leverage hyperbolic space’s inherent strength in modeling hierarchical data for the "coarse-to-fine" aspect, we introduce a hyperbolic network architecture and loss function. For the "few-shot" aspect, we improve model generalization by generating augmented samples based on hyperbolic probability distribution estimation. Experimental results validate the effectiveness of these improvements. 
\paragraph{Limitations.}
The performance of hyperbolic models is highly sensitive to curvature values, which are currently selected empirically without theoretical grounding or dynamic optimization strategies. Additionally, under few-shot conditions, prototypes estimated via probability distribution exhibit deviations from true distributions, leading to discrepancies between generated and real samples. While meta-learning methods could refine prototype estimation, such optimizations have not yet been explored in this work.

\bibliographystyle{abbrvnat}

\bibliography{ref}


\appendix

\section{Appendix}

\subsection{Overview of Riemannian geometry}

Riemannian geometry is a branch of differential geometry that focuses on smooth manifolds endowed with Riemannian metrics (\cite{Riemannian}).

\paragraph{Manifold}

A $d$-dimensional manifold ${\cal M}$ is a topological space which, for any point $m \in {\cal M}$, there exists a neighborhood $U \subset {\cal M}$ that is homeomorphic to $d$-dimensional Euclidean space. In other words, a manifold is a topological space that locally 'Euclidean', serving as a higher-dimensional generalization of curves and surfaces.

\paragraph{Tangent Space}
The tangent space is defined for every point $m \in {\cal M}$. It serves as a first-order linear approximation of the manifold at point $m$ , representing the local linearization of the manifold. Formally, it is the vector space composed of all vectors tangent to ${\cal M}$ at $m$ , analogous to the tangent line of a curve or the tangent plane of a surface in higher dimensions, which is denoted as ${{\cal T}_m}{\cal M}$.

\paragraph{Local Coordinate System}
A local coordinate system is a homeomorphic mapping from a local region of a manifold to Euclidean space. For every point $p$ on the manifold, the existence of such a mapping is guaranteed by the definition of a manifold. For 
$p \in {\cal M}$ and its neighborhood $U \subset {\cal M}$, which is homeomorphic to $d$-dimensional Euclidean space, we define a coordinate map \[\phi : U \to \mathbb{R}^d, \phi (p) = (x^1 (p), x^2 (p), \ldots, x^n (p)),\] mapping points on the manifold to Euclidean coordinates $\left( {{x^1},{x^2}, \ldots ,{x^n}} \right)$, where each $x^i : U \rightarrow \mathbb{R}$ is a smooth function from the manifold to real numbers. For example, on a 2-sphere (a 2-dimensional manifold), the latitude-longitude coordinate system serves as a local coordinate system, mapping points on the sphere (excluding the poles and the 180° meridian) to two real numbers (latitude and longitude), thereby achieving a local Euclidean representation of the manifold.

\paragraph{Riemannian Metric}
A Riemannian metric is a second-order covariant tensor field defined on a manifold ${\cal M}$ , satisfying smoothness, symmetry, and positive definiteness. It endows the tangent space at each point with an inner product structure, represented by a symmetric positive-definite matrix $G\left( x \right) = \left[ {{g_{ij}}\left( x \right)} \right]$ that varies smoothly with $x$. In a local coordinate system $\left( {{x^1},{x^2}, \ldots ,{x^n}} \right)$
, the metric can be expressed as \[{\rm{d}}{s^2}{\rm{ = }}\sum\limits_{i = 1}^n {\sum\limits_{j = 1}^n {{g_{ij}}\left( x \right){\rm{d}}{x_i}{\rm{d}}{x_j}} }.\] For example, in Euclidean space, the Riemannian metric is globally the identity matrix, corresponding to ${\rm{d}}{s^2}{\rm{ = d}}x_1^2 + {\rm{d}}x_2^2 +  \ldots  + {\rm{d}}x_n^2$ in local coordinates. Using the Riemannian metric, one can further define geometric quantities such as inner products, distances, and curvatures.

\paragraph{Norm and Inner Product}
Through the Riemannian metric, for a point $m$ on the manifold, the inner product on its tangent space ${{\cal T}_m}{\cal M}$ is defined as ${\left\langle {{\bf{u}},{\bf{v}}} \right\rangle _m} = {{\bf{u}}^T}G\left( m \right){\bf{v}}$ , where $G\left( m \right) $ is the Riemannian metric tensor at $m$. The norm of a tangent vector $v \in {{\cal T}_m}{\cal M}$ is similarly given by ${\left\| {\bf{v}} \right\|_m} = \sqrt {{{\left\langle {{\bf{v}},{\bf{v}}} \right\rangle }_m}}$.

\paragraph{Geodesics}
A geodesic generalizes the concept of uniform linear motion in Euclidean space to Riemannian manifolds. Formally, it is a curve on the manifold whose tangent vector has a covariant derivative that vanishes along the curve. For a curve $\gamma :\left[ {0,1} \right] \to {\cal M}$ connecting two points 
${\bf{x}},{\bf{y}}$ on the manifold, it is called a geodesic if it minimizes the energy function:\[L\left( \gamma  \right) = {\int_0^1 {\left\| {\gamma '\left( t \right)} \right\|} _{\gamma \left( t \right)}}{\rm{d}}t\] subject to boundary conditions $\gamma \left( 0 \right) = {\bf{x}}$ and $\gamma \left( 1 \right) = {\bf{y}}$. The Riemannian distance between ${\bf{x}},{\bf{y}}$ is then defined as:\[{\rm{dist}}\left( {{\bf{x}},{\bf{y}}} \right) = \min L\left( \gamma  \right),\gamma \left( 0 \right) = {\bf{x}},\gamma \left( 1 \right) = {\bf{y}}\]

\paragraph{Exponential and Logarithmic Maps}
The exponential map ${\exp _{\bf{x}}}:{{\cal T}_{\bf{x}}}{\cal M} \to {\cal M} $ generalizes vector addition in Euclidean space.  It describes how a point on the manifold is obtained by moving along a geodesic in a specified direction for a distance related to the norm, which is encoded by a tangent vector in the tangent space.Formally, for ${\bf{x}} \in {\cal M}$ and ${\bf{v}} \in {{\cal T}_{\bf{x}}}{\cal M}$, ${\exp _{\bf{x}}}\left( {\bf{v}} \right)$ is the endpoint $\gamma \left( 1 \right)$ of the geodesic $\gamma \left( t \right)$ starting at $x$ with initial velocity $v$. Specifically, $\gamma$ solves:\[{\gamma ^ * } = \mathop {\arg \min }\limits_\gamma  L\left( \gamma  \right),\gamma \left( 0 \right) = {\bf{x}},\gamma '\left( 0 \right) = {\bf{v}}\] and ${\exp _{\bf{x}}}\left( {\bf{v}} \right) = {\gamma ^ * }\left( 1 \right)$.The logarithmic map ${\log _{\bf{x}}}:{\cal M} \to {{\cal T}_{\bf{x}}}{\cal M}$ is defined as the inverse of the exponential map.

\paragraph{Curvature}
For Riemannian manifolds, there are various definitions of curvature. In this paper, the term "curvature" refers specifically to sectional curvature, which generalizes the Gaussian curvature of surfaces.

Given a Riemannian metric $ G\left( x \right) = \left[ {{g_{ij}}\left( x \right)} \right]$ and its corresponding local coordinate system $ \left( {{x^1},{x^2}, \ldots ,{x^n}} \right)$ , let ${G^{ - 1}}\left( x \right) = \left[ {{g^{ij}}\left( x \right)} \right]$ denote the inverse matrix. The Christoffel symbols of the second kind are defined as\[\Gamma _{ab}^c = \frac{1}{2}\mathop {\mathop \sum \limits^n }\limits_{k = 1} {g^{ck}}\left( {\frac{{\partial {g_{bk}}}}{{\partial {x^a}}} + \frac{{\partial {g_{ak}}}}{{\partial {x^b}}} - \frac{{\partial {g_{ab}}}}{{\partial {x^k}}}} \right),\] and the Riemann curvature tensor is given by\[R_{abc}^d = \frac{{\partial \Gamma _{ac}^d}}{{\partial {x^b}}} - \frac{{\partial \Gamma _{ac}^d}}{{\partial {x^b}}} + \mathop {\mathop \sum \limits^n }\limits_{k = 1} \left( {\Gamma _{ca}^k\Gamma _{bk}^d - \Gamma _{cb}^k\Gamma _{ak}^d} \right).\]Lowering the index, we obtain ${R_{abcd}} = \sum\limits_{k = 1}^n {{g_{dk}}R_{abc}^k} $. For a point $p \in {\cal M}$ and two linearly independent basis vectors ${\bf{u}},{\bf{v}}$ of ${{\cal T}_p}{\cal M}$, the sectional curvature $K$ at $p$ is computed as:\[K = \frac{{{{\left\langle {R\left( {{\bf{u}},{\bf{v}}} \right){\bf{v}},{\bf{u}}} \right\rangle }_p}}}{{\left\| {\bf{u}} \right\|_p^2\left\| {\bf{v}} \right\|_p^2 - \left\langle {{\bf{u}},{\bf{v}}} \right\rangle _p^2}} = \sum\limits_{i,j,k,l = 1}^n {\frac{{{R_{ijkl}}{{\bf{u}}^i}{{\bf{v}}^j}{{\bf{u}}^k}{{\bf{v}}^l}}}{{\left\| {\bf{u}} \right\|_p^2\left\| {\bf{v}} \right\|_p^2 - \left\langle {{\bf{u}},{\bf{v}}} \right\rangle _p^2}}} .\]

\begin{figure}[htbp]
  \centering
  \includegraphics[width=0.75 \textwidth]{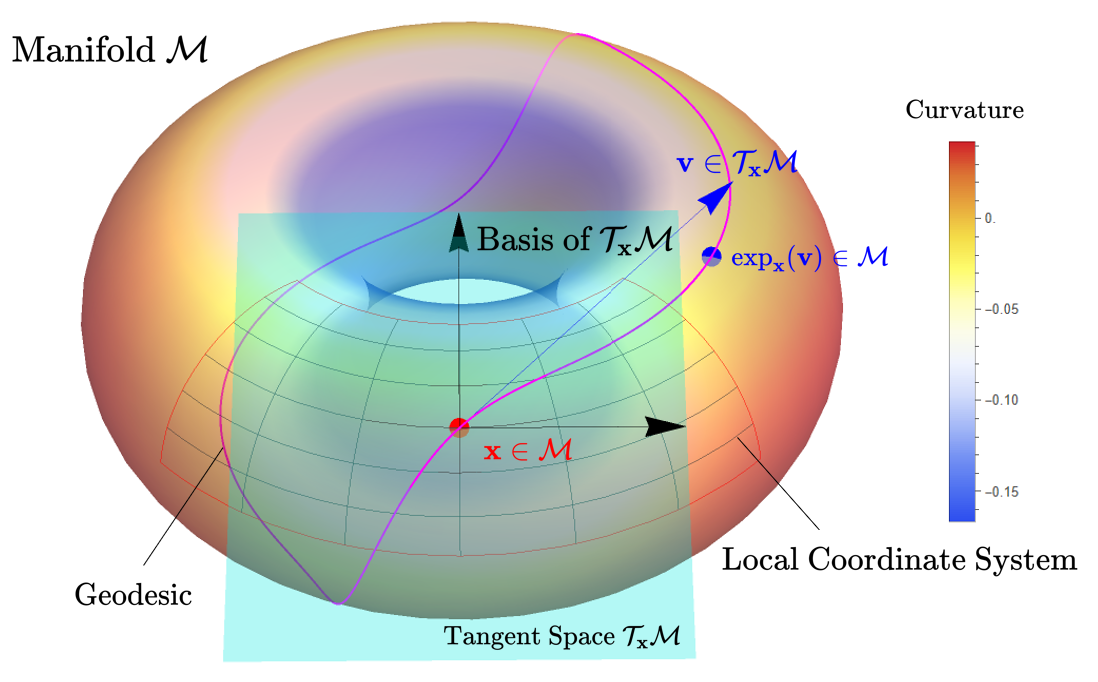}
  \caption{A schematic diagram illustrating key concepts in Riemannian geometry, depicting the tangent space, local coordinate system, geodesic passing through a point, exponential map, and the concept of curvature on a 2-dimensional torus manifold.}
\end{figure}

\subsection{The Poincaré ball model}

A Riemannian space is called a constant curvature space if the sectional curvature is identical at all points. Depending on the sign of the curvature $K$, such spaces can be classified into spherical space $\mathbb{S}^n (K > 0)$, Euclidean space $\mathbb{R}^n (K = 0)$ , and hyperbolic space $\mathbb{H}^n (K < 0)$. For machine learning tasks, spherical spaces are suitable for embedding data with ring-like structures, while hyperbolic spaces are better suited for hierarchical data (e.g., tree-structured data). This paper focuses on coarse-to-fine few-shot incremental learning tasks, which emphasize hierarchical data representations, and thus centers on hyperbolic space models and their corresponding machine learning methods.

Several models exist for representing hyperbolic spaces, including the Lorentz model, Klein-Beltrami model, and Poincaré ball model. Some of these models fail to converge to Euclidean space as $K \rightarrow 0$, potentially leading to poor network convergence. Therefore, this work adopts the Poincaré ball model, which avoids this issue. In the following sections, unless otherwise specified, $\left\| {\cdot} \right\|$ denotes the Euclidean norm of vectors, $\left\langle { \cdot , \cdot } \right\rangle$ denotes the Euclidean inner product of vectors.

The Poincaré ball model is a Riemannian space with the following Riemannian metric:\[{G_H}\left( {\bf{x}} \right) = {\left( {\lambda _{\bf{x}}^c} \right)^2}{G_E}\left( {\bf{x}} \right) = {\left( {\frac{2}{{1 - c{{\left\| {\bf{x}} \right\|}^2}}}} \right)^2}{I_n},\] where the curvature is a constant $ - c\left( {c > 0} \right)$, $\lambda _{\bf{x}}^c = \frac{2}{{1 - c{{\left\| {\bf{x}} \right\|}^2}}}$, and ${G_E}\left( {\bf{x}} \right)$ is the Euclidean metric (identity matrix). All vectors in the Poincaré ball must satisfy $\left\| {\bf{x}} \right\| < \frac{1}{{\sqrt c }}$, i.e., within the $n$-dimensional ball with radius $\frac{1}{{\sqrt c }}$: \[{\cal B}_c^n = \left\{ {{\bf{v}}\left| {{\bf{v}} \in {^n},\left\| {\bf{v}} \right\| < \frac{1}{{\sqrt c }}} \right.} \right\}.\]

Based on Riemannian geometry, for two points ${\bf{x}},{\bf{y}}$ in the Poincaré ball, the hyperbolic distance can be calculated as \[d_p^c({\bf{x}},{\bf{y}}) = \frac{1}{{\sqrt c }}{\cosh ^{ - 1}}\left( {1 + \frac{{2c{{\left\| {{\bf{x}} - {\bf{y}}} \right\|}^2}}}{{(1 - c{{\left\| {\bf{x}} \right\|}^2})(1 - c{{\left\| {\bf{y}} \right\|}^2})}}} \right),\]For a point ${\bf{w}}$ in the Poincaré ball and a tangent vector ${\bf{v}}$ in its tangent space, the exponential map can be calculated as \[\exp _{\bf{w}}^c({\bf{v}}) = {\bf{w}}{ \oplus _c}\left( {\tanh \left( {\sqrt c \frac{{\lambda _{\bf{w}}^c\left\| {\bf{v}} \right\|}}{2}} \right)\frac{{\bf{v}}}{{\sqrt c \left\| {\bf{v}} \right\|}}} \right)\]\\ The logarithmic map (inverse map) for points ${\bf{w}}$ and ${\bf{v}}$ can be calculated as \[\log _{\bf{w}}^c({\bf{v}}) = \frac{2}{{\sqrt c \lambda _{\bf{w}}^c}}{\tanh ^{ - 1}}\left( {\sqrt c \left\| {( - {\bf{w}}){ \oplus _c}{\bf{v}}} \right\|} \right)\frac{{( - {\bf{w}}){ \oplus _c}{\bf{v}}}}{{\left\| {( - {\bf{w}}){ \oplus _c}{\bf{v}}} \right\|}},\]where the operator ${ \oplus _c}$, which is called as Möbius addition, acts on vectors in hyperbolic space and preserves hyperbolic distance. For any points 
${\bf{u}},{\bf{v}},{\bf{w}}$ in the Poincaré ball, the property $d_p^c\left( {{\bf{u}},{\bf{u}}{ \oplus _c}{\bf{w}}} \right) = d_p^c\left( {{\bf{v}},{\bf{v}}{ \oplus _c}{\bf{w}}} \right)$ holds. The Möbius addition is computed as \[{\bf{x}}{ \oplus _c}{\bf{y}} = \frac{{(1 + 2c\left\langle {{\bf{x}},{\bf{y}}} \right\rangle  + c{{\left\| {\bf{y}} \right\|}^2}){\bf{x}} + (1 - c{{\left\| {\bf{x}} \right\|}^2}){\bf{y}}}}{{1 + 2c\left\langle {{\bf{x}},{\bf{y}}} \right\rangle  + {c^2}{{\left\| {\bf{x}} \right\|}^2}{{\left\| {\bf{y}} \right\|}^2}}}.\]

Specifically, as the curvature $c \rightarrow 0$, the Möbius addition degenerates to standard vector addition, the exponential map to vector addition, and the logarithmic map to vector subtraction, namely $\mathop {\lim }\limits_{c \to {0^ + }} \left( {{\bf{x}}{ \oplus _c}{\bf{y}}} \right) = {\bf{x}} + {\bf{y}}$, $\mathop {\lim }\limits_{c \to {0^ + }} \exp _{\bf{w}}^c({\bf{v}}) = {\bf{v}} + {\bf{w}}$, $\mathop {\lim }\limits_{c \to {0^ + }} \log _{\bf{w}}^c({\bf{v}}) = {\bf{v}} - {\bf{w}}$.Thus, the Poincaré ball model converges to Euclidean space as $c \rightarrow 0$.

\subsubsection{Optimization In Hyperbolic Space}

In Euclidean space, gradient descent is commonly employed to optimize loss functions. Assuming the loss function is $L$, the learning rate is $\eta$, and the current parameter is ${x_k}$ , the parameter is updated iteratively as:\[{x_{k + 1}} = {x_k} - \eta  \cdot \nabla L\left( {{x_k}} \right),\]However, in a general Riemannian space, this iterative formula requires adjustment via the Riemannian metric $G\left(  \cdot  \right)$, i.e.,\[{x_{k + 1}} = {x_k} - \eta  \cdot {G^{ - 1}}\left( {{x_k}} \right) \cdot \nabla L\left( {{x_k}} \right).\]Therefore, incorporating the Riemannian metric of the Poincaré ball model, the gradient descent iterative formula in this model should be:\[{x_{k + 1}} = {x_k} - \eta  \cdot \frac{{{{\left( {1 - c{{\left\| {{x_k}} \right\|}^2}} \right)}^2}}}{4} \cdot \nabla L\left( {{x_k}} \right).\]

\subsubsection{Process of round-off error}

Due to the exponential growth of the distance metric near the boundary of the Poincaré ball, when input vectors are very close to the surface of the Poincaré ball, minor computational errors can significantly affect the actual distance metric, thereby impacting model convergence (\cite{Numerical}). The following discusses the influence of rounding errors on the model.
Consider a vector ${\mathbf{x}} \in \mathcal{B}_c^n$ in the Poincaré ball model. If its norm satisfies $\left\| {\mathbf{x}} \right\| = \frac{1}{{\sqrt c }} - \varepsilon $, where $\varepsilon  > 0$ is a small positive number, the distance from the origin is given by $d_p^c({\mathbf{x}},0) = \frac{1}{{\sqrt c }}{\cosh ^{ - 1}}\left( {\frac{{1 + c{{\left\| {\mathbf{x}} \right\|}^2}}}{{1 - c{{\left\| {\mathbf{x}} \right\|}^2}}}} \right).$ Treating this as a function of $\left\| x \right\|$, consider the expansion near $x = \frac{1}{{\sqrt c }}$: \begin{align*}
  d_p^c({\mathbf{x}},0) & =  - \ln \left( {1 - \sqrt c \left\| {\mathbf{x}} \right\|} \right) + \ln 2 + O\left( {\left\| {\mathbf{x}} \right\| - \frac{1}{{\sqrt c }}} \right) \hfill \\
  & =  - \ln \left( {\sqrt c \varepsilon } \right) + \ln 2 + O\left( \varepsilon  \right) \hfill \\
  & =  \ln \frac{1}{\varepsilon } + \ln \frac{2}{{\sqrt c }} + O\left( \varepsilon  \right) \hfill
\end{align*}

Given that floating-point numbers have only 16-digit precision, let $\varepsilon  = {10^{-16}}$. At this point:\[d_p^c({\mathbf{x}},0) \approx 16\ln 10 + \ln \frac{2}{{\sqrt c }} \approx 37.5345 - \frac{1}{2}\ln c.\]
If the norm of the original Euclidean space vector exceeds this value, it will be mapped entirely to the boundary due to rounding errors after projection into hyperbolic space, causing the distance metric to become infinity in subsequent calculations and resulting in NaN losses and model divergence. Therefore, after completing hyperbolic space-related vector computations, additional post-processing is required for the output results. For example, the direction of the vector is preserved, but its norm is constrained to be within $\frac{1}{{\sqrt c }} - {10^{ - 6}}$. Vectors with norms exceeding this value are scaled down to this threshold to avoid model divergence caused by rounding errors.

\subsubsection{Figures illustrating concepts related to hyperbolic space}

\end{document}